\documentclass[letter, 10pt, conference]{ieeeconf}  
\usepackage{times}
\usepackage{epsfig}
\usepackage{graphicx}
\usepackage{amsmath}
\usepackage{amssymb}
\usepackage{pifont}
\usepackage{booktabs}
\usepackage{array}
\usepackage{caption}
\usepackage{tikz}
\usepackage{dsfont}
\usepackage[textwidth=2.2cm]{todonotes}
\usepackage{siunitx}
\usepackage[table]{}
\usepackage[dvipsnames]{}
\usepackage{xcolor,colortbl}
\let\labelindent\relax
\usepackage{enumitem}
\definecolor{falured}{rgb}{0.5, 0.09, 0.09}
\definecolor{teal}{rgb}{0.1176, 0.5019, 0.5019}
\newcommand{\xmark}{\ding{55}}%
\usepackage[colorlinks]{hyperref}
\ifdefined\nohyperref\else\ifdefined\hypersetup
  \definecolor{mydarkblue}{rgb}{0,0.08,0.45}
  \hypersetup{ %
    pdftitle={},
    pdfauthor={},
    pdfsubject={},
    pdfkeywords={},
    pdfborder=0 0 0,
    pdfpagemode=UseNone,
    colorlinks=true,
    linkcolor=mydarkblue,
    citecolor=mydarkblue,
    filecolor=mydarkblue,
    urlcolor=magenta,
    pdfview=FitH}

\usepackage[backend=biber,
            hyperref=true,
            url=false,
            isbn=false,
            doi=false,
            backref=false,
            style=ieee,
            citestyle=numeric-comp,
            sorting=nyt,
            block=none]{biblatex}

\IEEEoverridecommandlockouts                              % This command is only needed if 
\newcommand{\ctwotitleShort}{FSD}

\title{\LARGE \bf \ctwotitleShort:
Fast Self-Supervised Single RGB-D to Categorical 3D Objects}

\author{Mayank Lunayach$^1$\hspace{.2cm} Sergey Zakharov$^2$\hspace{.2cm} Dian Chen$^2$\hspace{.2cm} Rares Ambrus$^2$\hspace{.2cm} Zsolt Kira$^1$\hspace{.2cm} Muhammad Zubair Irshad$^1$\\[0.01cm]
$^1$Georgia Institute of Technology\hspace{.4cm} $^2$Toyota Research Institute\\[0.01cm]
{\tt\footnotesize \{lunayach, mirshad7, zkira\}@gatech.edu, \{firstname.lastname\}@tri.global}
}

\iffalse
    \newcommand{\sz}[1]{}
    \newcommand{\szn}[1]{}
    \newcommand{\zi}[1]{}
    \newcommand{\ml}[1]{}
\else
    \newcommand{\sz}[1]{{\color{red}{[#1]}}}
    \newcommand{\szn}[1]{{\color{red}{[#1]}}}
    \newcommand{\zi}[1]{{\color{magenta}{[#1]}}}
    \newcommand{\ml}[1]{{\color{orange}{[#1]}}}
\fi
    
\begin{document}

%%%%%%%%%%%%%%%

% \maketitle

\twocolumn[{%
\renewcommand\twocolumn[1][]{#1}%
\maketitle

\begin{center}
    \centering
    \vspace{-0.4cm}
    \includegraphics[width=0.95\textwidth]{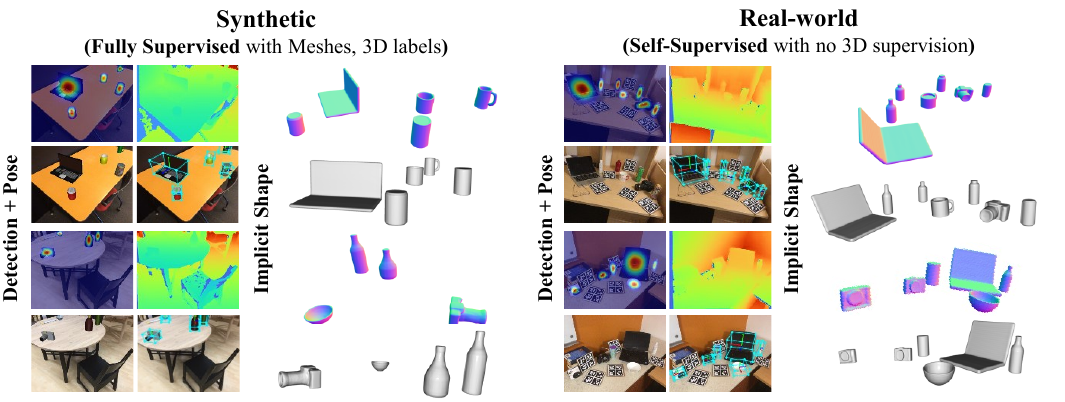}
    % \captionsetup{width=\linewidth}
    \captionof{figure}{
    \textbf{Overview:} We present~\textbf{\ctwotitleShort}, a fast self-supervised categorical 6D pose and size estimation and shape reconstruction framework. Our method is a fully feed-forward approach that doesn't require any real-world 3D labels such as meshes or 6D pose annotations and it does not necessitate inference time optimization. 
    Our results demonstrate effective sim-to-real transfer on the NOCS~\cite{wang2019normalized} real-world test-set.
    }
   \label{fig:teaser}
\end{center}%
}]

\thispagestyle{empty}
\pagestyle{empty}

\begin{abstract}

In this work, we address the challenging task of 3D object recognition without the reliance on real-world 3D labeled data. Our goal is to predict the 3D shape, size, and 6D pose of objects within a single RGB-D image, operating at the category level and eliminating the need for CAD models during inference. While existing self-supervised methods have made strides in this field, they often suffer from inefficiencies arising from non-end-to-end processing, reliance on separate models for different object categories, and slow surface extraction during the training of implicit reconstruction models; thus hindering both the speed and real-world applicability of the 3D recognition process. Our proposed method leverages a multi-stage training pipeline, designed to efficiently transfer synthetic performance to the real-world domain. This approach is achieved through a combination of 2D and 3D supervised losses during the synthetic domain training, followed by the incorporation of 2D supervised and 3D self-supervised losses on real-world data in two additional learning stages. By adopting this comprehensive strategy, our method successfully overcomes the aforementioned limitations and outperforms existing self-supervised 6D pose and size estimation baselines on the NOCS test-set with a 16.4\% absolute improvement in mAP for 6D pose estimation while running in near real-time at 5 Hz. Project page: \href{https://fsd6d.github.io/}{fsd6d.github.io}

\end{abstract}

\section{INTRODUCTION}

\looseness=-1
The task of 3D perception from monocular RGB-D input has garnered significant attention within the domains of computer vision~\cite{He2017,zhou2019objects,gkioxari2019mesh, heppert2023carto} and robotics~\cite{jiang2021synergies,ferrari1992planning,laskey2021simnet,irshad2022centersnap}. This compelling problem holds immense relevance, with wide-ranging implications in critical areas such as autonomous navigation~\cite{irshad2022sasra,9561806} and robotic manipulation~\cite{cifuentes2016probabilistic,jiang2021synergies,laskey2021simnet,irshad2022centersnap}. At its core, this challenge revolves around the extraction of intricate 3D object representations from a single RGB-D perspective of the environment. This entails the precise estimation of a multitude of object attributes, encompassing 3D shape, 6D pose (comprising orientation and position), size, and visual appearance of individual object instances. Due to its inherent complexity, the problem is ill-posed, and predicting 3D information from 2D images can be ambiguous.

\begin{table*}[t]
    \centering
    
    \newcommand{\yes}{\ding{52}}
    \newcommand{\no}{\ding{55}}

\resizebox{\textwidth}{!}{
\begin{tabular}{l c c c c c c} \hline
    \textbf{Feature} & 
    NOCS~\cite{wang2019normalized} &
    ShAPO~\cite{Irshad2022ShAPOIR} & RePoNet~\cite{Fu2022CategoryLevel6O} & CenterSnap~\cite{irshad2022centersnap} & SSC-
    6D~\cite{peng2022self} &
    \cellcolor{yellow!40}\textbf{\ctwotitleShort~(Ours}) \\
    \hline
    Without real-world 3D labels & \textcolor{falured}{\xmark} & \textcolor{falured}{\xmark} & \textcolor{teal}{\checkmark} & \textcolor{falured}{\xmark} & \textcolor{teal}{\checkmark} & \cellcolor{yellow!40}\textcolor{teal}{\checkmark} \\
    3D shape & \textcolor{falured}{\xmark} & \textcolor{teal}{\checkmark} & \textcolor{falured}{\xmark} & \textcolor{teal}{\checkmark} & \textcolor{teal}{\checkmark} & \cellcolor{yellow!40}\textcolor{teal}{\checkmark} \\
    One model for all categories & \textcolor{falured}{\xmark} & \textcolor{teal}{\checkmark} & \textcolor{falured}{\xmark} & \textcolor{teal}{\checkmark} & \textcolor{falured}{\xmark} & \cellcolor{yellow!40}\textcolor{teal}{\checkmark} \\
    Simultaneous detection and 6D Pose & \textcolor{falured}{\xmark} & \textcolor{teal}{\checkmark} & \textcolor{falured}{\xmark} & \textcolor{teal}{\checkmark} & \textcolor{falured}{\xmark} & \cellcolor{yellow!40}\textcolor{teal}{\checkmark} \\
    Without post-optimization & \textcolor{teal}{\checkmark} & \textcolor{falured}{\xmark} & \textcolor{teal}{\checkmark} & \textcolor{teal}{\checkmark} & \textcolor{teal}{\checkmark} &\cellcolor{yellow!40}\textcolor{teal}{\checkmark} \\
    \hline

\end{tabular}
}
\caption{\textbf{Comparison with other 6D pose estimation and shape reconstruction methods:}~\textcolor{teal}{\checkmark} indicates a method has the feature, and \textcolor{falured}{\xmark} indicates that it doesn't. 
    }
\label{tab:intro_comparison}
\end{table*}

\looseness=-1
Various data-driven (supervised learning) approaches have been proposed for this challenging problem~\cite{Irshad2022ShAPOIR, sundermeyer2020augmented, niemeyer2020differentiable, groueix2018, kuo2020mask2cad}. However, annotating 3D data can be quite expensive. Furthermore, when it comes to making robots capable of transitioning seamlessly from simulated environments to the real world, labeling real-world data may not be always feasible. In this context, the appeal of 2D supervision becomes evident as a cost-effective and widely accessible alternative. 2D supervision comes at minimal cost, and pseudo labels generated by recent methods like Segment Anything~\cite{kirillov2023segany} and MiDaS~\cite{Ranftl2022} for in-the-wild data are of very high quality.

However, the absence of methods to generate robust 3D real labels necessitates the exploration of innovative training paradigms, particularly those grounded in self-supervised learning, to address this constraint. Recent developments have introduced self-supervised approaches that operate without the dependency on 3D labels~\cite{peng2022self, Fu2022CategoryLevel6O, Fu2022CategoryLevel6O}. While these proposed methods have exhibited promise, they may not be feasible for practical applications. For example, object detection and 3D prediction happen in separate stages~\cite{peng2022self, Fu2022CategoryLevel6O, Fu2022CategoryLevel6O}, resulting in sub-optimal inference times. Additionally, the linear growth of model size with the number of object categories poses scalability challenges. Consequently, the applicability of such approaches in real-world settings may be limited. Notably,~\cite{Fu2022CategoryLevel6O} needs additional unlabelled real data to enable effective transfer to real-world scenarios.

\looseness=-1
To account for the discussed limitations, we propose a novel self-supervised end-to-end method to infer the 3D shape, size, and 6D poses of multiple objects. We employ a multi-stage training strategy to transfer synthetic performance to real-world domains. Firstly, the model is pre-trained on synthetic data (CAMERA~\cite{wang2019normalized}) using both 2D and 3D supervision (which comes for free). This is followed by joint training on synthetic and real data (Real 275~\cite{wang2019normalized}). During this phase, only 2D labels from the real data are employed. In the final stage, the model undergoes fine-tuning exclusively on the real data, once again relying solely on 2D labels from the real dataset. To account for no direct 3D supervision from real data, we employ 3D self-supervision using chamfer loss. Specifically, we get a pseudo ground truth for point clouds by back-projecting the input depth maps into 3D space. These estimated point clouds serve as a loose supervision mechanism for guiding the predicted point clouds. As a result, our approach consistently yields superior quantitative and qualitative results when compared to robust baseline methods.

Our proposed method is most similar to~\cite{Irshad2022ShAPOIR} and~\cite{peng2022self} but with key differences. Unlike our method,~\cite{Irshad2022ShAPOIR} 
relies on 3D labeled real-world data and uses post-optimization after training (resulting in less efficient processing times). Conversely, in contrast to~\cite{peng2022self}, our method is end-to-end~(detection and 3D prediction happen in one forward pass) and utilizes a single universal model for all categories. Key distinctions from existing methods are concisely summarised in Table~\ref{tab:intro_comparison}. Our contributions are summarized as follows:
\begin{itemize}
    \item A novel~\textbf{multi-stage training pipeline} for~\textbf{fast and efficient} shape reconstruction and 6D pose and size estimation~\textbf{without requiring real-world 3D labels}.
    
    \item Our approach achieves state-of-the-art results and outperforms existing baselines for self-supervised 6D pose estimation, showing~\textbf{over 16\% absolute improvement} in mAP for 6D pose at 10\textdegree \SI{10}{\cm} on the NOCS~\cite{wang2019normalized} test set.
 
    \item A~\textbf{faster-batched shape extraction} and an~\textbf{end-to-end feed-forward approach} making fine-tuning pipeline and model inference~\textbf{orders of magnitude faster} then competing baselines.
\end{itemize}

\section{RELATED WORK}
% Our method learns to infer the 3D shape, 6D pose, and size of multiple objects in an image without using 3D labeled data, that is, doing self-supervised object-centric scene reconstruction from single-view RGB-D. It relates to areas like object understanding, pose estimation, scene reconstruction, and self-supervised learning.
% \noindent\textbf{Object reconstruction:} Monocular 3D reconstruction has made significant strides, yielding diverse output representations such as pointclouds, voxels, or meshes~\cite{park2019deepsdf,mescheder2019occupancy,chen2019learning}. DVR~\cite{niemeyer2020differentiable} and SRN~\cite{sitzmann2019scene} extract information from multiple views, employing differentiable rendering and ray-marching routines, respectively. Recently, many NeRF~\cite{mildenhall2020nerf, Barron2023ZipNeRFAG, Barron2021MipNeRF3U, Barron2021MipNeRFAM} based methods have also been proposed. NeRFs~\cite{mildenhall2020nerf} perform color and density regression along a ray and employ volumetric rendering to generate pixel colors. However, it's worth noting that NeRF-based approaches~\cite{li2020neural, Niemeyer2020GIRAFFE, Ost_2021_CVPR} frequently exhibit a tendency to overfit to specific scenes, limiting their generalizability. Further, they tend to require dense viewpoint annotations. Our method, however, is able to get the 3D representations of instances unseen during the training and learns from a set of view-point independent single-view RGBD inputs.
\noindent\textbf{Implicit shape representations:}
Scalar field approximators have emerged as a prominent direction to model 3D shapes. Notable works include Occ-Net~\cite{mescheder2019occupancy}, IM-Net~\cite{chen2019learning} and DeepSDF~\cite{park2019deepsdf}. These methods output either an SDF (Signed Distance Field) or an occupancy estimate for every 3D coordinate.  NGLOD~\cite{takikawa2021neural}, ROAD~\cite{zakharov2022road}, and MeshSDF~\cite{remelli2020meshsdf} have been proposed for efficient octree representation and differentiable mesh representation, respectively. Our method utilizes implicit fields along with a fast octree-based sampling~(Section~\ref{sec:diff_shape}) to decode shapes which is used for self-supervised loss during training.

\noindent\textbf{6D pose and size estimation:} Work streams using pose regression~\cite{irshad2022centersnap, kehl2017ssd, wang2019densefusion, xiang2018posecnn}, template matching~\cite{kehl2016deep, sundermeyer2018implicit, tejani2014latent} and establishing correspondences~\cite{wang2019normalized, dpodv2, park2019pix2pose,hodan2020epos, goodwin2022} have been proposed. Most works, however, focus only on pose estimation and not the simultaneous 3D shape prediction. We deal with the task of end-to-end 6D pose and size estimation without relying on any test-time optimization or real-world 3D labels.
\begin{figure*}[ht!]
   \centering
    \includegraphics[width=1.0\linewidth]{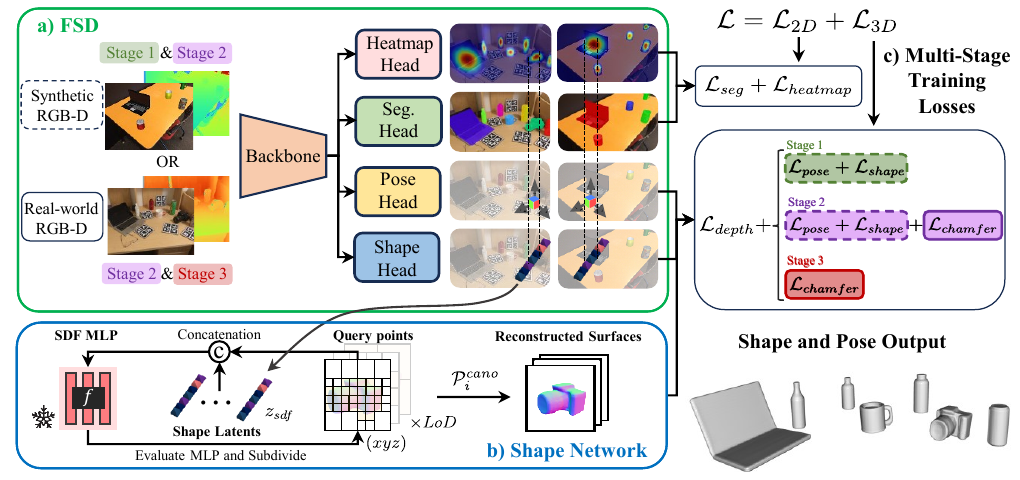}
    \caption{
    \textbf{Method:} 
      a) Forward pass of the proposed model across different training stages. From a single-view RGB-D observation, the model predicts a segmentation mask, depth map, object heatmap, pose map, and shape map.
      b) Batchified recursive point sampling is illustrated where the batch of concatenated 3D points and latent vectors are evaluated for SDF using a frozen shape auto-decoder.
      c) Losses for different training stages. Losses are color-coded based on the training stages and have solid out lines for real data and dotted lines for synthetic data. 
    }
   \label{fig:method}
\end{figure*}

\noindent\textbf{Self-supervised methods for 6D pose estimation:} DSC-PoseNet~\cite{Yang2021DSCPoseNetL6} and Self6d~\cite{Wang2020Self6DSM} have introduced instance-level 6D pose estimation methods that heavily rely on the rendering of CAD models of the target objects. However, such approach may face scalability challenges when applied to real-world scenarios. In contrast,~\cite{Park2019LatentFusionED} proposed a pose estimation technique that utilizes multiple reference images rather than 3D scans, which may also be not practical when applied to real-world data at scale. In contrast, both~\cite{He2022TowardsSC} and~\cite{You2022CPPFTR} carried out sim2real adaption after training on synthetic data.  For a more robust generalization, SSC-6D~\cite{peng2022self}, RePoNet~\cite{Fu2022CategoryLevel6O}, and CPS++\cite{Manhardt2020CPSIC} train on both real and synthetic data in combination.
\vspace{-1em}

% can also discuss the 3D humans work here

\section{METHOD}
We propose \ctwotitleShort, a novel learning-based object-centric scene understanding method. From an RGB-D observation, it estimates 6D pose, 3D shape, and size of all the seen object instances in an image without requiring real-world 3D labels to train its method. It performs object detection, localization, and 3D reconstruction. All three steps happen end-to-end without any post-processing. The network design is inspired by~\cite{Irshad2022ShAPOIR} and consists of two major components: a) A detection module for detecting the 2D locations of the objects; b) A 3D prediction module for predicting the 3D shape, 6D pose, and size of the objects. For our synthetic 3D priors, we use DeepSDF~\cite{park2019deepsdf} which is a learned continuous Signed Distance Function (SDF) representing shapes of different categories. We first train it offline on ShapeNet~\cite{chang2015shapenet} to get a shape decoder. Then the decoder is frozen and is used to recover the implicit object shape using predicted latent codes and query points contained in a unit cube.

Concretely, given a single-view RGB-D observation containing an image $I \in \mathds{R}^{h \times w \times 3}$ and depth map $D \in \mathds{R}^{h \times w}$, our method,~\textbf{\ctwotitleShort{}}, infers the 6D pose $\tilde{\mathbf{\mathcal P}}$ $\in$ $SE(3)$, 1D scale $\tilde{\mathcal{T}} \in \mathds{R}^1$, and 3D shape~(as $SDF$) for each detected object. The framework is depicted in Fig.~\ref{fig:method}. First, an FPN-based backbone network extracts and fuses features from the RGB-D input (following \cite{Irshad2022ShAPOIR}), which are processed by task-specific heads to get dense predictions of heatmap, shape, pose, and category score. Next, we describe our network's forward pass in Sections~\ref{sec:seg} and Section~\ref{sec:pose}, our fast batched surface extraction in~\ref{sec:diff_shape} and lastly our novel multi-stage 3D training strategy in~\ref{sec:3dlearning} along with self-supervised loss in~\ref{sec:chamfer} to aid the effective transfer of fully supervised synthetic domain learning to real-world domain without requiring any real 3D labels.

\subsection{Segmentation and Heatmap head}
\label{sec:seg}
The segmentation head and heatmap head predict the categories and 2D centers for objects in the scene. For $n$ categories, the segmentation head predicts an $(n+1)$-channel logit map (an extra channel for the background class) $\hat{M}\in \mathds{R}^{\frac{h}{R} \times \frac{w}{R}\times 1}$, which is supervised with ground truth instance masks. The object 2D centers are predicted as heatmaps~$\hat{H}\in [0,1]^{\frac{h}{R} \times \frac{w}{R}\times 1}$ by the heatmap head, where each local maxima in the heatmap ~$\hat{H}$ becomes the detected point~$(\hat{x}_{i}, \hat{y}_{i})$. L2 loss $\mathcal{L}_{heatmap}=\sum_{xy}\left(\hat{H}- H\right)^{2}$ is applied to supervise the head where ground truth heatmaps $H$ are constructed from 2D coordinates with Gaussian kernels. $R$ is the downsampling factor of the backbone.

\subsection{Pose, Shape and Depth head}
\label{sec:pose}
  In parallel, the pose and shape heads predict a pose map~${z_{pose}}$ $\in$ $\mathds{R}^{\frac{h}{R} \times \frac{w}{R} \times 13}$ and a shape map~${z_{shape}}$ $\in$ $\mathds{R}^{\frac{h}{R} \times \frac{w}{R} \times D}$. $D=64$ is the dimension of the shape latent embedding. For each instance, shape and pose embeddings are queried using the predicted object centers from the heatmap $\hat{H}$ as follows:
\begin{equation}
    {z_{sdf}}_{i} = z_{shape}[x_{i}, y_{i}, :]; \quad {sRT}_{i} = z_{pose}[x_{i}, y_{i}, :]
\end{equation}

\noindent where ($x_{i}, y_{i}$) is one of $N$ predicted object centers in $\hat{H}$, and $N$ is the number of instances detected by the model in the given input. These two heads are supervised during training stage 1 and 2 (see Section ~\ref{sec:3dlearning}) using synthetic data, where L1 losses of pose and shape embedding are calculated between the ground truth and estimated values, weighted by the estimated heatmap $\hat{H}$.

Latent code ${z_{sdf}}_{i}$ for each instance is fed to a SDF (Signed Distance Function) based auto-decoder to get the predicted point cloud in the canonical space $\mathcal{P}^{cano}_{i}$ as done in~\cite{park2019deepsdf}. 

Shape estimation using SDFs is computationally expensive. Thus, we employ an octree-based point sampling method, similar to~\cite{Irshad2022ShAPOIR}. We implement a novel batched differentiable shape extraction and discuss it in Section~\ref{sec:diff_shape}. The shape decoder is pre-trained and frozen as shown in the section b of Fig.~\ref{fig:method}. The point cloud in the camera frame is estimated as $\mathcal{P}^{cam}_{i}=\left({sRT}_{i}\right)\mathcal{P}^{cano}_{i}$.
Additionally, we supervise a depth head with an auxiliary depth loss. Since the real-world depth maps are noisy with artifacts, we introduce the same noise in synthetic data and try to recover the clean depth. This enhances the performance transfer from synthetic to real.

\subsection{Differentiable Surface Extraction}
\label{sec:diff_shape}

When provided with query points $q_i$ and their corresponding signed distance values $s_i$, we require a differentiable approach to access the implicit surface. Merely selecting query points based on distance values does not allow us to compute derivatives with respect to the latent vector ${z_{sdf}}_{i}$. However, by computing signed distance functions (SDFs) with respect to their locations, we can efficiently calculate the normal vector at each surface point during a reverse pass $n_i = \frac{\partial f(q_i; \mathbf{z})}{\partial q_i}$ similar to~\cite{Irshad2022ShAPOIR, zakharov2020autolabeling}.

Normals point toward the nearest surface, and signed distance values provide precise distance information, enabling us to map the query location to a 3D surface position denoted as $p_i$:
\begin{equation}
p_i = q_i - \frac{\partial f(q_i;{z_{sdf}}_{i})}{\partial q_i} f(q_i;{z_{sdf}}_{i}).
\end{equation}
\paragraph{Batchified Recursive Point Sampling}
To efficiently extract surface points for predicted objects associated with their respective latent vectors, we employ a recursive point extraction method inspired by~\cite{Irshad2022ShAPOIR}. We start by defining a coarse voxel grid and computing SDF values for each point using our trained SDF network. Voxels with SDF values exceeding the voxel grid size for the current resolution level are discarded, while the remaining voxels undergo subdivision, generating eight new voxels in each iteration. This process continues until reaching the desired resolution level, initiated at the level of detail (LoD) 1 and progressing to LoD 6. Once completed, we extract the point coordinates, along with their corresponding SDF values and normals, projecting them onto the object surface using the previously described iso-surface projection procedure.

To efficiently recover shapes for multiple objects concurrently, we implement a batchified version of the extraction algorithm. Since different objects may have distinct shapes, we traverse a single octree structure encompassing all predicted objects. This involves initializing a coarse grid for each predicted latent vector and collectively traversing them while monitoring boundaries that separate points belonging to different objects. Upon reaching the final LoD, we extract point clouds for each object based on established boundaries.

\subsection{Chamfer loss}
\label{sec:chamfer}
To provide a 3D learning signal for the real data, we compute the Chamfer loss between the estimated point cloud from scale, pose, and SDF and the ground-truth point cloud. For each instance, similar to \cite{peng2022self}, a point cloud of the visible points $\mathcal{P}^{real}_{i}$ is lifted from the input depth map $D$ using the camera intrinsics $K$; correspondingly, the estimated point cloud is calculated from $\mathcal{P}^{cam}_{i}=\left({sRT}_{i}\right)\mathcal{P}^{cano}_{i}$. Typical chamfer loss calculates distance between all pairs of points the two point clouds. However, the depth-lifted point clouds are noisy and prone to outliers. Therefore, to make it robust to noises, we use a thresholded version of chamfer loss and only calculate the loss if two points are less than $\epsilon$ units apart. Specifically,
\vspace{-0.05cm}
\begin{align*}
    \mathcal{L}_{chamfer} = \frac{1}{N_{p}} \sum_{p_j \in \mathcal{P}^{real}_{i}} \max \left(0, \epsilon - \min_{p_{k} \in \mathcal{P}^{cam}_{i}} \|p_j - p_{k}\|_2 \right)
    \\
    + \frac{1}{N_{p}} \sum_{p_j \in \mathcal{P}^{cam}_{i}} \max \left(0, \epsilon - \min_{p_{k} \in \mathcal{P}^{real}_{i}} \|p_j - p_{k}\|_2 \right)
\end{align*}
\noindent where $\epsilon > 0$ and $N_{p}$ is the total number of points satisfying $ \|p_{j} - p_{k}\|_2 < \epsilon, ~\forall~p_j \in \mathcal{P}^{real}_{i}~\text{and}~\forall~p_{k}~\in \mathcal{P}^{cam}_{i}$. To stabilize the convergence, $\mathcal{P}^{cano}_{i}$ is detached from the gradient computation graph. 

\subsection{3D learning strategy}
\label{sec:3dlearning}
In this section, we describe our approach to address the challenge of training a 3D reasoning model with the absence of real 3D labels. Our strategy consists of three distinct stages: pre-training on synthetic data followed by mixed training with a combination of synthetic and real data, and lastly fine-tuning on real data. We employ this sequence of training steps to effectively enable 3D reasoning and generalization on real-world data.
\subsubsection{Stage 1: Pre-training on Synthetic Data}
In the initial stage, we perform pre-training using synthetic data from the CAMERA dataset~\cite{wang2019normalized}. Synthetic data inherently provides 3D labels, allowing us to leverage this information to learn 3D priors. This pre-training stage serves as a foundational step in our approach. Formally, the objective looks like the following: $    \mathcal{L}_{pretrain} = \mathcal{L}_{seg} + \mathcal{L}_{depth} +
 \mathcal{L}_{heatmap} + \mathcal{L}_{pose} +  \mathcal{L}_{shape}$
\begin{table*}[ht!]
    \centering
    \scriptsize
    \newcommand{\yes}{\ding{52}}
    \newcommand{\no}{\ding{55}}
    
\resizebox{\textwidth}{!}{
\begin{tabular}{l c c c c c c c} \hline
    Method & Type & \textbf{IOU25}~$\uparrow$ & \textbf{IOU50}~$\uparrow$ & \textbf{5\textdegree \SI{5}{\cm}}~$\uparrow$ & \textbf{5\textdegree \SI{10}{\cm}}~$\uparrow$ & \textbf{10\textdegree \SI{5}{\cm}}~$\uparrow$ & \textbf{10\textdegree \SI{10}{\cm}}~$\uparrow$ \\
    \hline
    NOCS~\cite{wang2019normalized} & & \textbf{84.8} & 78.0 & 10.0 & 9.8 & 25.2 & 25.8 \\
    Metric Scale~\cite{lee2021category} &  & 81.6 & 68.1 & 5.3 & 5.5 & 24.7 & 26.5\\
    ShapePrior~\cite{tian2020shape} & Supervised &  81.2 & 77.3 & 21.4 & 21.4 & 54.1 & 54.1\\
    CASS~\cite{chen2020learning} &  &  84.2 & 77.7 & 23.5 & 23.8 & 58.0 & 58.3\\
    CenterSnap~\cite{irshad2022centersnap} &  & 83.5 & \textbf{80.2} & \textbf{27.2} & \textbf{29.2} & \textbf{58.8} & \textbf{64.4}\\
    \hline
    CPS++~\cite{Manhardt2020CPSIC} &  & - & 17.7 & - & - & $\leq 22.3$ & - \\
    SSC-6D~\cite{peng2022self} & Self-supervised & \textbf{83.2} & 73.0 & 19.6 & - & 54.5 & 56.2\\
    \textbf{\ctwotitleShort{} }(Ours) &  & 80.9 & \textbf{77.4} & \textbf{28.1} & \textbf{34.4} & \textbf{61.5} & \textbf{72.6}\\
    \hline
\end{tabular}
}
    \caption{\textbf{Results on REAL275:} Quantitative comparison of our proposed method with competing baselines. Supervised denotes utilizing both camera and real-world 3D supervision during training whereas self-supervised methods do not utilize any real-world 3D annotations. All methods train on both CAMERA25 and Real275 datasets. }
    \label{tab:results_cam}
\end{table*}

\subsubsection{Stage 2: Mixed Training}
Directly transitioning from synthetic pre-training to fine-tuning on real data can lead to two undesirable effects: 1) the forgetting of 3D priors due to the absence of explicit 3D supervision in real data, and 2) overfitting on the real training data without meaningful 3D learning. To address these challenges, we introduce a mixed training phase following pre-training. During mixed training, each batch of inputs comprises a combination of real and synthetic data. While synthetic data points come with 3D labels, real data points do not. This design ensures that the model retains the 3D priors acquired during pre-training while adapting to the nuances of real-world data. We conceptualize this intermediate mixed-training step as a form of soft exposure to real-world data.

For our experiments, we keep the ratio of synthetic to real samples as 5. Specifically, for a sample $b$ in a mixed batch $B$, total loss looks like this, 
\vspace{-0.1cm}
\begin{align*}
    \mathcal{L}_{mixed} = \mathcal{L}_{seg} + \mathcal{L}_{depth} +
 \mathcal{L}_{heatmap} + \\ \mathds{1}(b \in syn) ( \mathcal{L}_{pose} +  \mathcal{L}_{shape}) + \mathds{1}(b \in real) \mathcal{L}_{chamfer} 
\end{align*}
where $\mathds{1}(b \in syn)$ denotes if the sample $b$ is synthetic and $\mathds{1}(b \in real)$ denotes if the sample $b$ is real. To better facilitate learning from two different data distributions, batch normalization with fixed mean and variance (learnt during pre-training) is used.
\subsubsection{Stage 3: Fine-tuning on Real Data}
In the final stage, we focus on maximizing the learning from relevant real data. Fine-tuning is carried out exclusively on the real data, allowing the model to refine its understanding of real-world 3D structures. The following loss objective is used: $    \mathcal{L}_{finetune} = \mathcal{L}_{seg} + \mathcal{L}_{depth} +
 \mathcal{L}_{heatmap} + \mathcal{L}_{chamfer} $

All loss terms are weighted using loss weights $\lambda$s that are omitted above for clarity.

\section{EXPERIMENTS}
\subsection{Datasets}

We leverage both real and synthetic datasets. \textbf{CAMERA} \cite{wang2019normalized} has 275K synthetic images with 3D annotations. The training dataset includes 1085 object models spanning across 6 categories: \textit{bottle}, \textit{bowl}, \textit{camera}, \textit{laptop}, and \textit{mug}. The evaluation dataset has 184 object models spanning across the same set of categories. \textbf{REAL} \cite{wang2019normalized} train set has 7 scenes spanned across 4300 images, with test-set having 6 scenes spanning across 2750 images. 
\subsection{Implementation details}
 Weights of $\mathcal{L}_{seg}$, $\mathcal{L}_{depth}$, $\mathcal{L}_{heatmap}$, $\mathcal{L}_{pose}$, $\mathcal{L}_{shape}$, and $\mathcal{L}_{chamfer}$ are 1, 1, 100, 0.1, 0.1 and 10 respectively, determined empirically. The distance threshold ($\epsilon$) for $\mathcal{L}_{chamfer}$ is set at 0.2. Batch size of 32 is used and multi-gpu training on 7 NVIDIA A100s is performed. Adam~\cite{Kingma2015AdamAM} optimizer with 0.9 momentum is used with the learning rate of $6 \times 10^{-4}$ decaying with polynomial decay~(exponent=0.9) across epochs until convergence. Color jitter augmentation is used for both synthetic and real data. Flip augmentation is only used for real data because it is non-trivial to flip 3D labels for synthetic. 

 Predicting rotations in an unconstrained setting such as ours can be challenging. From a network perspective, they are arbitrary 9 real numbers~($SE3$ representation), and the constructed rotation metrics may not be orthogonal, which may introduce unwanted shear transformations. To control this effect,~\cite{Levinson2020AnAO} introduced an SVD-based technique to ensure the orthogonalization of the predicted rotation metrics. We use this orthogonalization during our training. For training the SDF-based shape decoder, an MLP with 8 layers and a hidden size of 512 is used and is trained offline for 2000 epochs. PyTorch~\cite{NEURIPS2019_9015} and PyTorch3D~\cite{ravi2020pytorch3d} is used for implementation.

\subsection{Metrics}
\looseness=-1
Following ~\cite{irshad2022centersnap, Irshad2022ShAPOIR, Fu2022CategoryLevel6O, peng2022self}, the performance of 3D object detection and 6D pose estimation is evaluated independently. For 3D object detection, we report the average precision for IoU 25 and IOU 50 thresholds. For 6D pose estimation, average precision for which error is less than $x$\textdegree~for rotation and $y$ cm for translation is reported where $(x, y) \in \{5, 10\}$.

\subsection{Baselines}
We compare against both fully-supervised and self-supervised baselines. For a fair comparison, we compare against methods that do not perform post-processing or post-optimization \cite{Irshad2022ShAPOIR} or use any additional unlabelled data for a better synthetic to real world transfer \cite{Fu2022CategoryLevel6O}.
\subsubsection{Fully-supervised baselines} 

\begin{enumerate}[label=(\alph*)]

\item  \textbf{NOCS}\cite{wang2019normalized}: This architecture extends the Mask-RCNN framework to predict the NOCS map and utilizes a similarity transform with depth information for pose and size prediction.

\item \textbf{Shape Prior}\cite{tian2020shape}: This method infers a 2D bounding box for each object and predicts shape deformations.

\item  \textbf{CASS}\cite{chen2020learning}: It employs a two-stage approach. First, it detects 2D bounding boxes, and then it regresses the pose and size of the objects.

\item \textbf{Metric-Scale}\cite{lee2021category}: An extension of NOCS, this method predicts the object center and metric shape separately.

\item \textbf{CenterSnap}\cite{irshad2022centersnap}: It is an end-to-end method that does object detection and 3D prediction in one forward pass.
\end{enumerate}

\subsubsection{Self-supervised baselines}
\begin{enumerate}[label=(\alph*)]
    \item \textbf{CPS++}~\cite{Manhardt2020CPSIC}: This approach proposed coarse 3D alignment of 3D centroids before doing fine alignment using chamfer loss for self-supervision. 
    \item \textbf{SSC-6D}~\cite{peng2022self}: It does 3D prediction on center cropped images having the individual detected objects. It uses a two-stage pipeline (warm up on the synthetic data followed by mixed training) and depth lifted point clouds for self-supervision.
\end{enumerate}

\renewcommand{\arraystretch}{1.0}

\begin{table}[t!]
    \centering
\resizebox{0.48\textwidth}{!}{%
\begin{tabular}{l c c c c} \hline
    Method & IOU25 & IOU50 & 5\textdegree \SI{10}{\cm} & 10\textdegree \SI{10}{\cm} \\
    \hline
    PT &  27.3 & 26.1 & 6.0 & 34.7\\
    PT + MT & 70.9 & 67.3 & 13.8 & 51.5\\
    PT + FT & \textbf{82.1} & \textbf{78.4} & 16.9 & 57.2\\
    PT + FT + FT & 81.4 & 75.7 & 14.2 & 54.1\\
     \hline
    PT + MT + FT (Ours) & 80.9 & 77.4 & \textbf{34.4} & \textbf{72.6}\\
    \hline
\end{tabular}
}
    \caption{\textbf{Ablation:} Effect of training strategy (results on NOCS Real 275). PT, MT and FT stand for Pre-training, Mixed-training and Fine-tuning respectively.}
    \vspace{-0.3cm}\label{tab:ablation_training_type}
\end{table}
\begin{table}[b!]
\vspace{-8pt}
\begin{minipage}{.48\columnwidth}
\centering
\resizebox{0.99\columnwidth}{!}{
\begin{tabular}{@{}lc@{}}
\toprule
Method       & \textbf{Inference time}  \\ 
\midrule
SSC-6D~\cite{peng2022self}      & 1.99s  \\
Ours & \textbf{0.20s}\\
\bottomrule
\end{tabular}
}
\captionof{table}{Quantitative inference-time comparison per image on A100 GPU}
\label{tab:inference_time}
\end{minipage}
\centering
\hfill\begin{minipage}{.48\columnwidth}
\centering
\resizebox{0.99\columnwidth}{!}{
\begin{tabular}{@{}lc@{}}
\toprule
Method & \textbf{Shape time} \\ 
\midrule
DeepSDF~\cite{park2019deepsdf} & 0.15s  \\
Octree-SDF~\cite{Irshad2022ShAPOIR} & 0.11s\\
Batched Octree~(Ours) & \textbf{0.01s}\\
\bottomrule
\end{tabular}
}
\captionof{table}{{Quantitative shape extraction time with LoD 6 for rows 2, 3 and resolution 64 for~\cite{park2019deepsdf}}}
\label{tab:shape_time}
\end{minipage}
\vspace{-16pt}
\end{table}
\renewcommand{\arraystretch}{1.0}
\begin{table}[t!]
    \scriptsize
    \centering
\resizebox{0.48\textwidth}{!}{%
\begin{tabular}{l c c c c c c c} \hline
    Method & & & & & & IOU50 & 10\textdegree \SI{10}{\cm} \\
    \hline
    Zero-shot & & & & & & 26.1 &  34.7\\
    2D supervision & & & & & & 4.2 &  4.1\\
     \hline
 Ours & & & & & & \textbf{77.4} &  \textbf{72.6}\\
    \hline
\end{tabular}
}
    \caption{\textbf{Ablation:} Effect of loss objectives used during training stages (results on NOCS Real 275)}
    \vspace{-0.5cm}
    \label{tab:ablation_losses}
\end{table}

\begin{figure}[t!]
\centering
\includegraphics[width=0.93\columnwidth]{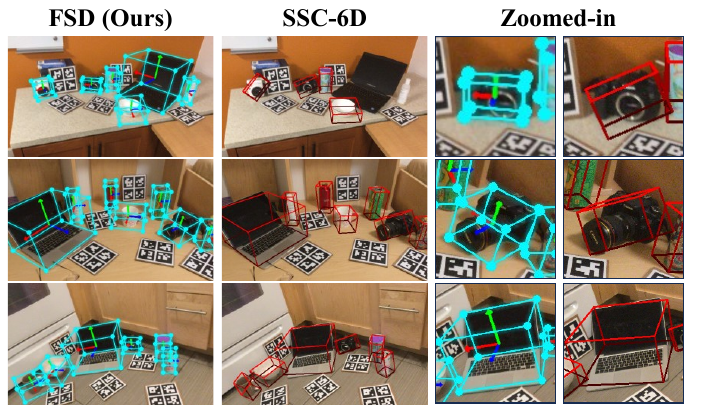}
\captionof{figure}{
  \textbf{Qualitative comparison:} Our method vs. SSC-6D~\cite{peng2022self} on NOCS Real275 test-set. 
  }  

\label{qualitative}
\vspace{-0.5cm}
\end{figure} 

\subsection{Comparison with the baselines}
\subsubsection{3D object detection and 6D pose estimation}
\looseness=-1
Results are tabulated in the Table \ref{tab:results_cam}. Against \textbf{fully-supervised baselines}, we observe superior performance (absolute gains as high as 8.2\%) in the pose estimation performance across all the mAP metrics and very similar 3D detection performance (IoU). This is a significant result as all the supervised baselines used 3D real labels, unlike our method. This hints that our method was able to learn useful 3D inductive biases without being directly trained for it. Self-supervised methods are known to generalize better. Even though supervised methods have access to 3D labels, chamfer loss against noisy pointclouds helps our method in learning a more generalizable representation.    Against \textbf{self-supervised baselines}, significant performance gains are also observed. There's an improvement in IoU50 by absolute 4.4\% and consistent improvement in pose-estimation performance (absolute gains as high as 16.4\%). 

\subsubsection{Inference speeds}
We compare the inference speed of our model in time per image with the baseline in the Table~\ref{tab:inference_time}. Our method being single-shot and end-to-end makes it faster. It is also worth noting that SSC-6D~\cite{peng2022self} uses one model per category, thus allowing for more parameters for representation learning. Specifically, its space complexity is $\mathcal{O}(n_{categories})$ (where $n_{categories}$ is the number of categories in the dataset) whereas ours is $\mathcal{O}(1)$.

\subsection{Qualitative comparison}
In Fig.~\ref{qualitative}, we compare the output of our proposed model with SSC-6D~\cite{peng2022self}. Given the RGB-D input, the model learns to estimate accurate 6D poses (visualized bounding boxes with orientations). Our model achieves near-perfect pose~(Fig. \ref{qualitative}) and shape estimations~(Fig. \ref{fig:teaser}) without being trained on 3D real data and having not seen the test-time instances during training (category-level estimation).

\subsection{Ablation}
We ablate various design choices for our proposed method and answer the following questions.
\subsubsection{Why not directly fine-tune the pre-trained model?}
\looseness=-1
Directly fine-tuning the pre-trained model may seem attractive but in doing so, model may forget the 3D priors it learned during the pre-training step. Mixed training allows the model to smoothly transition from 3D labeled data points to 3D unlabeled data points while learning about the real data. This has been demonstrated in Table~\ref{tab:ablation_training_type} where training for three stages and mixed training is superior than other combinations. 

\subsubsection{How important is chamfer loss?}
A case may be made to fine-tune only using the 2D losses on the real data. Doing so may not help the model in learning 3D attributes of real data. Only using 2D losses (i.e. no chamfer loss) for real data in our training strategy makes it forget the 3D learnings. Zero-shot performance is directly evaluating the pre-trained model.  Results are summarized in the Table~\ref{tab:ablation_losses}.

\subsubsection{How fast is batched shape extraction?}
We introduce a novel batched shape extraction by batch querying Signed Distance Functions using Octree based sampling. This is summarised in the Table~\ref{tab:shape_time}. Faster shape extraction directly leads to faster training speeds and thus making it suitable for training on large-scale in-the-wild datasets.

\section{CONCLUSION}
In this paper, we introduce \ctwotitleShort, a novel method for estimating the 6D pose, size, and shape of objects in a scene. Our method 1) is end-to-end (object detection, localization, and 3D predictions happen in one forward pass), 2) uses one universal model for all the categories and does not need a separate model for different categories, 3) does not use 3D real labels 4) is faster to train compared to baselines. We propose a novel multi-stage training strategy to maximize learning from unlabelled real data. To our knowledge, ours is the first work to attempt all these goals together. Our method also achieved a significant increase in the pose estimation while outperforming strong self-supervised pose estimation baselines. We hope the utility and performance of our proposed method will motivate similar works in the future. 
% \input{chapters/chapter2_files/future_work}

%%%%%%%%%%%%%%%%%%%%%%%%%%%%%%%%%%%%%%%%%%%%%%%%%%%%%%%%%%%%%%%%%%%%%%%%%%%%%%%%
% \bibliographystyle{IEEEtran}
% \bibliography{ieeeconf/egbib}

\begin{flushright}
\printbibliography
\end{flushright}

\end{document}